\documentclass[10pt,twocolumn,letterpaper]{article}

\usepackage[pagenumbers]{cvpr} 

\usepackage{multirow}

\usepackage{enumitem}
\usepackage{booktabs}
\usepackage{multirow}
\usepackage{multicol}
\usepackage{makecell}
\usepackage[dvipsnames]{xcolor}
\usepackage{array}

\usepackage[dvipsnames]{xcolor}

\definecolor{cvprblue}{rgb}{0.21,0.49,0.74}
\usepackage[pagebackref,breaklinks,colorlinks,citecolor=cvprblue]{hyperref}

\def\paperID{10094} 
\def\confName{CVPR}
\def\confYear{2024}

\newcommand{\fancyname}{\textit{GeoDeformer}}

\title{\fancyname: Geometric Deformable Transformer for Action Recognition}

\author{
Jinhui Ye$^1$, Jiaming Zhou$^{1}$, Hui Xiong$^{1,2,3,\dagger}$, Junwei Liang$^{1,2,\dagger}$  \\
$^1$Thrust of Artificial Intelligence, HKUST (Guangzhou), Guangzhou, China\\
$^2$Department of Computer Science and Engineering, HKUST, Hong Kong SAR, China\\
$^3$ HKUST Fok Ying Tung Research Institute, Guangzhou, China  \\
\normalsize\tt \{jye624,jzhou760\}@connect.hkust-gz.edu.cn  \\
\normalsize\tt xionghui@ust.hk  \normalsize\tt junweiliang@hkust-gz.edu.cn \\
}

\begin{document}
\maketitle

\begin{figure*}[h!]
	\centering		\includegraphics[width=0.98\textwidth]{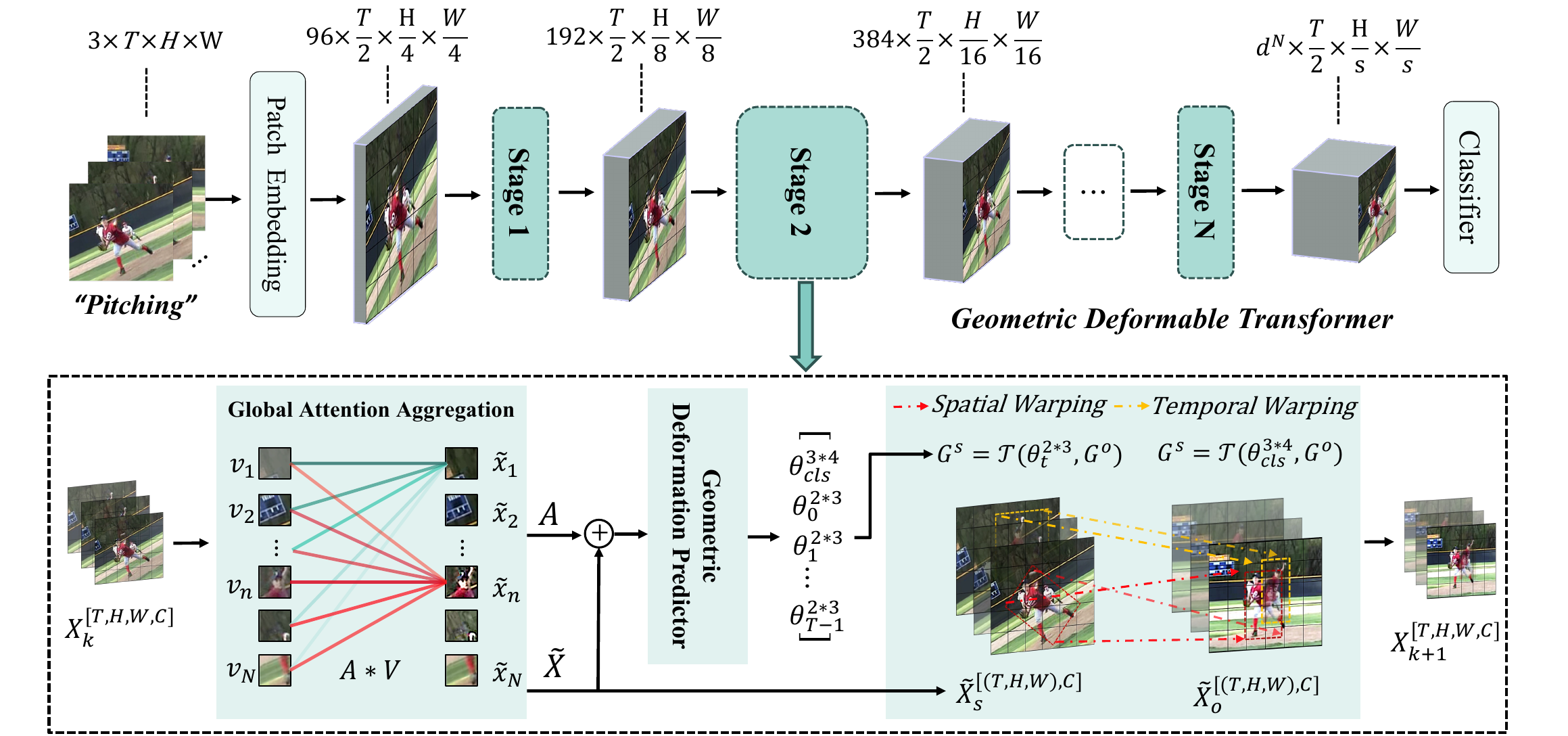} 
\caption{
     Overview of the Geometric Deformable Transformer (\fancyname)  framework for action recognition. One {\fancyname} block is structured into three main components: the Global Information Aggregation (\(\mathcal{A}\)) that captures contextual relationships across the video sequence, the Geometric Deformation Predictor (\(\mathcal{D}\)) that identifies necessary geometric changes within the video, and the Spatial-Temporal Warping Module (\(\mathcal{T}\)) that applies these deformations to enhance the feature representation.  
 }

\vspace{-4mm}

\label{fig:overview}
\end{figure*}

\begin{abstract}

Vision transformers have recently emerged as an effective alternative to convolutional networks for action recognition. 
However, vision transformers still struggle with geometric variations prevalent in video data. 
This paper proposes a novel approach, {\fancyname}, designed to identify and  eliminate the variations inherent in action video 
by integrating geometric comprehension directly into the ViT architecture. 
Specifically,  at the core of {\fancyname} is the Geometric Deformation Predictor, a module designed to identify and quantify potential spatial and temporal geometric deformations within the given video. 
Spatial deformations adjust the geometry within individual frames
, while temporal deformations capture the cross-frame geometric dynamics, reflecting motion and temporal progression. 
To demonstrate the effectiveness of our approach, 
we incorporate it into the established MViTv2 framework, replacing the standard self-attention blocks with {\fancyname} blocks.
Our experiments at UCF101, HMDB51, and Mini-K200 achieve significant increases in both Top-1 and Top-5 accuracy, establishing new state-of-the-art results with only a marginal increase in computational cost. 
Additionally, visualizations affirm that {\fancyname} effectively manifests explicit geometric deformations and minimizes geometric variations.
Codes and checkpoints will be released.

\end{abstract}   
\let\thefootnote\relax\footnotetext{$^\dagger$ Corresponding authors.}

\section{Introduction}
Recently, vision transformers (ViT)~\cite{dosovitskiy2020transformers} have quickly gained popularity in computer vision with their good performance and simple design. 
In the field of action recognition, ViT has rapidly emerged as a versatile and dominant backbone architecture~\cite{li2022mvitv2}.
The cornerstone of ViTs in handling visual data lies in their capability to capture longer-range dependencies. This is primarily achieved through the self-attention operation, a novel approach that facilitates analyzing and comparing individual token features extracted from disparate visual patches. 
However, unlike language data, visual data inherently possesses distinct geometric characteristics, such as spatial orientation, scale variance, and perspective differences. 
These geometric variations in present unique challenges to visual recognition tasks, especially in action recognition where capturing variations across space and time is crucial.
To address these distinct geometric characteristics in visual data, one common approach within the ViT framework is data augmentation. 
These methods enhance the model's adaptability to spatial and temporal variances through artificial modifications of training samples, altering spatial attributes such as scale, orientation, and perspective.

%


However, those data augmentation approaches predominantly rely on hand-crafted and pre-defined transformations to elicit variations from training samples. This approach can lead to increased computational costs and might prove ineffective in covering the full range of real-world geometric variances.
Moreover, it does not inherently equip ViTs with an understanding of the underlying geometric principles, which is vital for comprehensively interpreting non-rigid deformations, diverse viewpoints, and complex temporal dependencies in visual data.


To this end, we introduce {\fancyname}, a novel approach that learns to reduce the variations inherent in action video by geometric comprehension and transformations. 
We draw inspiration from the geometric transformation theory and 
the principles outlined in \cite{kosiorek2019stacked}, which suggest human vision's reliance on coordinate frames.
The key idea behind {\fancyname} is the incorporation of explicit geometric deformations into the ViT architecture.
Specifically, we design a module called Geometric Deformation Predictor ($\mathcal{D}$) to identify potential geometric deformations within the given video and quantify them into geometric transformation parameter ${\theta}s \in \mathcal{R}^{2 \times 3}$ or ${\mathcal{R}^{3 \times 4}}$. 
Then, we use a differentiable warping function $\mathcal{T}$ to execute spatial and temporal geometric deformations. This is achieved by resampling the feature map following the coordinate mappings behind of $\theta$s (see Eqn.~\ref{eqn:2D} and Eqn.~\ref{eqn:3d}).
Compared with the transformer attention mechanism which learns relations between visual tokens by global attention, 
the deformation predictor explicitly learns geometric transformations that reduce inherent variations in visual data.

To demonstrate the effectiveness of our approach, 
we incorporate it into the established MViTv2 framework~\cite{li2022mvitv2}, replacing the standard self-attention blocks with our proposed {\fancyname} blocks. 
As illustrated in Table~\ref{tab:main_results},  our {\fancyname}  significantly improves performance on three popular action recognition datasets, establishing the new state-of-the-art.  Notably, on the UCF101 dataset, {\fancyname} achieves a significant absolute increase of 4.1\% in Top-1 accuracy. Similarly, on the larger-scale Mini-K200 dataset, we observe an improvement of 1.15\% in Top-1 accuracy, with only a 4\% increase in computational cost.
Further, as visualized in Section~\ref{sec:visualize}, {\fancyname} effectively manifests explicit geometric deformations and reduces geometric variations.

To summarize, this paper makes the following key contributions:

(1) We design a novel 
geometric deformable transformer
block that can identify and quantify geometric variations in visual data. 
This block allows spatial-temporal deformations directly applied to the feature maps, enabling a transformative approach to processing and interpreting dynamic visual content.

(2) We pioneer the exploration of incorporating spatial-temporal deformations in visual transformer architectures for action recognition. This novel approach allows visual transformers to learn geometric invariant representations explicitly.

(3) We provide extensive experimental evidence across multiple datasets to demonstrate that the {\fancyname} consistently improves upon the performance of the ViT architectures in action recognition tasks, setting new state-of-the-art performances. Additionally, visualizations affirm that {\fancyname} effectively manifests explicit geometric deformations and minimizes geometric variations.

\section{Related Work}

\subsection{Vision Transformer Backbone}

Since the Vision Transformer (ViT)~\cite{dosovitskiy2020transformers} was adapted from natural language processing (NLP) to the visual domain, there has been a surge of innovations aimed at enhancing its applicability to visual tasks~\cite{yang2021focal, liu2022video, li2022uniformer}. These advancements have primarily concentrated on learning multi-scale features for dense prediction tasks and developing efficient attention mechanisms.
The MViT~\cite{li2022mvitv2} extends the ViT to effectively process multi-scale features, enhancing its feature extraction capabilities across scales.  Swin Transformer~\cite{liu2022video} modifies the traditional structure to a hierarchical model with shifted window-based self-attention, addressing scalability and improving efficiency in large-scale visual data processing.  DETR\cite{dai2021dynamic}, DAT~\cite{Xia_2022_CVPR}, and DVT~\cite{wang2022deformable} optimize transformer efficiency by localizing global attention to smaller key sets, conserving computational resources. Uniformer~\cite{li2022uniformer} capitalizes on the strengths of both CNNs and self-attention, offering a hybrid approach that leverages the best of both architectures. 
These developments mainly focus on techniques like windowed attention~\cite{dong2022cswin,liu2022video}, global tokens~\cite{chen2021regionvit, jaegle2021perceiver, sun2021visual}, focal attention~\cite{yang2021focal}, and dynamic token sizes~\cite{rao2021dynamicvit}, aimed at enhancing model efficiency and integrating additional inductive biases.

Despite these significant strides, a notable challenge remains: the limited ability of existing ViT models to effectively handle viewpoint changes and complex geometric variations in dynamic visual environments.
In this work, we are focusing on endowing the ViT with the capability to perform geometric deformations on visual data.


\subsection{Spatial and Temporal Alignment}

Effective spatial-temporal alignment is crucial in action recognition for capturing the nuanced interplay between spatial dynamics and temporal progression in video content~\cite{lin2017inverse, wang2022deformable}. 
Conventional CNNs offer significant insights, where their shared-weight architecture is inherently position-independent.
Another common strategy, data augmentation, aims to enhance adaptability to spatial and temporal variances by artificially modifying training samples, altering attributes such as scale, orientation, and perspective~\cite{liu2022video, li2022mvitv2, li2022uniformer}. 
However, this approach still falls short in explicitly addressing viewpoint changes and other complex spatial-temporal variations.
To address these limitations, contemporary approaches often expand feature representations either by integrating layers capable of extracting motion cues from raw RGB frames, like 3D convolutions~\cite{ji20123d, tran2015learning, tran2018closer, xie2018rethinking} and other temporal operators~\cite{hommos2018using, lin1811temporal,sun2018optical,wang2020video}, or by incorporating pre-extracted optical flow~\cite{feichtenhofer2016convolutional, wang2016temporal}.
However, these methods of expanding features tend to create complex, high-dimensional feature maps, leading to increased computational costs and a potential loss of geometric interpretability. 

Spatial Transformer Networks (STNs)~\cite{jaderberg2015spatial, lin2017inverse, ye2023spatial} are more related to our approach, which allows neural networks to perform spatial transformations like rotation and scaling on input images. However, STNs primarily focus on spatial manipulation in static images and do not fully address video temporal aspects. 
Our work extends these concepts by incorporating explicit geometric transformations for spatial and temporal video sequence variations.  
Additionally, we integrate these spatial-temporal geometric transformations into the Vision Transformer (ViT) framework, creating a more comprehensive and effective approach for action recognition.








\section{Methods}

In this section, we describe the proposed  {\fancyname} in detail.
We have seen the great success of Vision Transformers~\cite{dosovitskiy2020transformers} in the field of computer vision in general. 
Yet, there has not been a principled way of resolving geometric variations in the given data, especially in videos. 
The key idea behind {\fancyname} is the incorporation of explicit geometric deformations into the transformer architecture to reduce such variations.
In the following section, we first describe the background of geometric transformation~(\cref{sec:bg}). Next, we provide an overview of our {\fancyname} block~(\cref{sec:overview}), 
which updates the visual feature map with geometric deformations.
Then, we explain the vital designs of {\fancyname} for spatial-temporal deformation modeling, i.e., geometric deformation predictor $\mathcal{D}$~(\cref{sec:stdlm}) and spatial-temporal warping $\mathcal{T}$~(\cref{sec:warp}).
Finally, we conclude by detailing the construction of our video action recognition network~(\cref{sec:arch}).

\subsection{Preliminaries}
\label{sec:bg}

We first revisit the fundamental geometric transformations. Geometric transformations in image processing are encapsulated by a parameter matrix $ \theta $.
For example, it is represented as a $ 2 \times 3 $ matrix for affine transformation. 
This matrix defines a coordinates mapping, which resamples the source into the target image, facilitating transformations like scaling, rotation, and translation.

Formally, given an image $ \mathcal{I}_{in} \in \mathbb{R}^{H \times W \times C} $ with height $ H $, width $ W $, and channels $ C $, and a coordinates grid $ \mathcal{G} \in \mathbb{R}^{H \times W \times 2} $ consisting of pixel indices $ (x, y) $. 
The geometric transformations on the image are defined as resampling the input image $ \mathcal{I}_{in}$ into $ \mathcal{I}_{out} $ according to a coordinates mapping, which is represented as:

\vspace{-4mm}
\begin{align}
\label{eqn:2D}
    \mathcal{T}(\theta, \mathcal{G}): 
    \begin{bmatrix}
    X^s \\
    Y^s \\ 
    \end{bmatrix}
    =
    \begin{bmatrix}
    p_1 & p_2 & p_3\\
    p_4 & p_5 & p_6 \\ 
    \end{bmatrix}
    \begin{bmatrix}
    X^o \\
    Y^o \\
    1 \\
    \end{bmatrix}
\end{align}

where the $ [X^s, Y^s] $ denotes the source pixel locations that map to the target pixel $ [X^o, Y^o] $.
The pixel intensity values are transferred from the input image to the output image using the relationship:

\vspace{-4mm}
\begin{align}
\label{Egn：warp}
\mathcal{I}_{out}[X^o, Y^o] = \mathcal{I}_{in}[X^s, Y^s]
\end{align}


For a video $V$ represented by $\mathbb{R}^{T \times H \times W \times C}$,   temporal dimension $T$ is taken as $Z$ axis, and the coordinates mapping is represented as:
\vspace{-4mm}
\begin{align}
\label{eqn:3d}
    \mathcal{T}(\theta, \mathcal{G}):
    \begin{bmatrix}
    Z^s \\
    X^s \\
    Y^s \\
    \end{bmatrix}
    =
    \begin{bmatrix}
    p_1 & p_2 & p_3 & p_4\\
    p_5 & p_6 & p_7 & p_8\\
    p_9 & p_{10} & p_{11} & p_{12}\\
    \end{bmatrix}
    \begin{bmatrix}
    Z^o \\
    X^o \\
    Y^o \\
    1 \\
    \end{bmatrix}
\end{align}


\subsection{Geometric Deformable Transformer Block}
\label{sec:overview}

Our proposed Geometric Deformable Transformer (\fancyname) 
learns to reduce geometric variations in video data.
The $k$-th block is responsible for learning geometric transformation parameters and updating the feature map from $\tilde{X}_k$ into $X_{k+1}$ by resampling based on the learned coordinate mappings $\mathcal{T}(\theta, \mathcal{G})$. 
Figure \ref{fig:overview} depicts the forward processing of the {\fancyname} block, 
which consists of three key operations:




\begin{enumerate}
\item \textbf{Global Information Aggregation ($\mathcal{A}$)}: 
Given a video transformed into a visual token map $X_k$ via patch embedding,
this operation utilizes a basic self-attention mechanism to evaluate and encode the relationships throughout the entire feature map. 
The process not only updates individual patch tokens to encompass a broader context but also captures the global dependencies among these tokens, within them by an attention matrix $A$.

\vspace{-2mm}
\begin{align}
    \tilde{X}_{k}, A = \mathcal{A}(X_k) = \text{Self-Attention}(X_{k})
\label{eqn:A}
\end{align}

\item \textbf{Geometric Deformation Predictor ($\mathcal{D}$)}: This module detects and quantifies potential geometric deformations in the input video, encoding the necessary alterations to account for the spatial-temporal dynamics of the video content by geometric deformation parameters. 
Specifically, our {\fancyname} is designed to predict two types of transformation parameters.

\noindent \textbf{Frame-wise Deformation Parameters:} This deformation type is articulated through a sequence of transformation matrices \( \theta_{0:{T-1}} = [\theta_0, \theta_1, ..., \theta_{T-1}] \), where \(  \theta_t \in \mathbb{R}^{2 \times 3} \). 
Each matrix independently applies spatial transformations to its respective frame, conforming to Eqn.~\ref{eqn:2D}.

\noindent \textbf{Cross-frame Deformation Parameter:} 
Cross-frame deformation applies to the whole video using a single transformation matrix, \( \theta_{cls} \in \mathbb{R}^{3 \times 4} \), which operates on the temporal dimension as well. 

\vspace{-4mm}

\begin{align}
    \Theta = [\theta_{\text{cls}}, \theta_0, \theta_1, ..., \theta_{T-1}] = \mathcal{D}(\tilde{X}_{k}, A) 
\end{align}


\item \textbf{Spatial-Temporal Warping (\( \mathcal{T} \))}: This module utilizes the deformation parameters \( \Theta \) to perform a coordinate mapping transformation from \( \mathcal{G}_{in} \) to \( \mathcal{G}_{out} \) according to Eqn.~\ref{eqn:2D} and Eqn.~\ref{eqn:3d}. It effectively reshapes the visual patch token feature map, applying geometric alterations that complement the global attention mechanisms. 

\vspace{-4mm}
\begin{align}
    \mathcal{G}_{in} = \mathcal{T}(\Theta,  \mathcal{G}_{out}) \\
    X_{k+1}[\mathcal{G}_{out}] = \tilde{X}_{k}[\mathcal{G}_{in}]
\end{align}


\end{enumerate}

\subsection{Geometric Deformation Predictor}
\label{sec:stdlm}
The Geometric Deformation Predictor is tasked with identifying the latent geometric deformations within the given video and quantifying them into deformation parameters \( \Theta \). 
To detailed and accurately model both spatial and temporal geometric deformations within video sequences, ensuring a comprehensive spatial-temporal alignment,
$\mathcal{D}$ deftly manages spatial-temporal deformations in videos by harnessing self-attention and CNN components, delivering a unified representation of spatial and temporal changes. As depicted in Figure~\ref{fig:location}, the module's operation is detailed as follows:

\begin{figure}[ht]
	\centering		\includegraphics[width=0.40\textwidth]{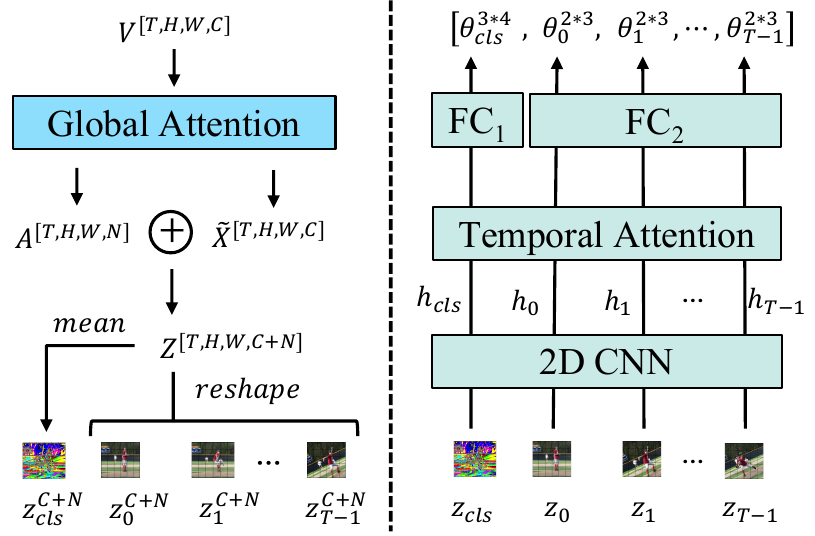} 
    \caption{Illustration of the Geometric Deformation  Predictor, $\mathcal{D}$. This diagram depicts the process of learning transformation parameters that account for both frame-wise and cross-frame deformations.  For detailed parameter settings, refer to Appendix~\ref{apd:hype_para}.}
\label{fig:location}
\end{figure}

\begin{enumerate}
 
    \item \textbf{Frame-wise Feature Construction}:
    According to ~\cref{sec:overview} and Eqn.~\ref{eqn:A}, Global Information Aggregation ($\mathcal{A}$) updates feature map $X_k$ to \( \tilde{X}_k \) using a basic self-attention mechanism. Additionally, the accompanying attention matrix \( A \in \mathbb{R}^{T \times H \times W \times N} \), with \( N \) representing the product of temporal and spatial dimensions, is employed to assess and encode the relationships throughout the entire feature map. This attention matrix sheds light on the complex interrelationships between pixels. 
    The subsequent fusion of \( \tilde{X_k} \) and \( A \) yields an integrated feature map \( Z \), which is dissected into per-frame features \( \{z_t\}_{t=0}^{T-1} \), each with an expanded channel dimension \( \mathbb{R}^{H \times W \times (C+N)} \). A global class token \( z_{cls} \), synthesizing the video's overarching story, is computed as the mean of these frame-level features.

    \vspace{-4mm}
    \begin{align}
    Z &= \text{concat}(\tilde{X}, A)  \\
    z_{cls} &= \text{mean}({z}_{t=0}^{T-1})
    \end{align}

    \item \textbf{Frame Feature Extraction}: This component harnesses spatially aware 2D CNN layers to distill spatial context from the composite feature map \( Z \), producing a series of frame embedding $h \in \mathbb{R}^C$. 

    \vspace{-4mm}
    \begin{align}
        [h_{cls}, h_0, h_1, ..., h_{T-1}] &= \text{CNN}(z_{cls} ,Z)
    \end{align}

    \item \textbf{Temporal Feature Integration}: 
    Even though frame-wise geometric deformations are applied independently to each frame, it is crucial to capture temporal interdependencies within the video sequence. This is achieved through a lightweight self-attention mechanism that processes the series of frame-level features \( [h_0, h_1, ..., h_{T-1}] \) along with a global class (\( cls \)) token \( h_{cls} \). This process synthesizes a comprehensive feature set \( H^* = [h^*_{cls}, h^*_0, h^*_1, ..., h^*_{T-1}] \), encapsulating both individual frame characteristics and the overall context of the video.

    \vspace{-4mm}
    \begin{align}
      H^{*} &= \text{temporal-attention}([h_{cls}, h_0, h_1, ..., h_{T-1}])
      \label{eqn:temporal}
    \end{align}

    \item \textbf{Transformation Parameter Generation}: The final stage employs a fully-connected layer to convert the extracted features into specific transformation parameters. This includes parameters for frame-wise geometric deformation \( \theta_{0:{T-1}} \)and cross-frame deformation \( \theta_{cls} \)  , enabling precise spatial-temporal manipulation.
    
    \vspace{-4mm}
    \begin{align}
        \theta_{cls} &= \text{FC}_1(h^{*}_{cls}) \in \mathbb{R}^{3 \times 4}  \\
        \theta_{0:{T-1}} &= \text{FC}_2([h^{*}_0:h^{*}_{T-1}]) \in \mathbb{R}^{2 \times 3}
    \end{align}
\end{enumerate}

\subsection{Spatial-Temporal Warping}
\label{sec:warp}

In this section, we introduce the differentiable warping function $ \mathcal{T} $. 
The warping function $\mathcal{T}$ defines the mapping between these coordinates, enabling the transition of features from the original to the target map through a process of differentiable spatial and temporal interpolation. It enables the reflection of the geometric transformations identified by the deformation network $\mathcal{D}$.
This function performs two types of warping, i.e., Spatial Warping and Temporal Warping, aligning features within and across video frames:

\begin{enumerate}
    \item \textbf{Spatial Warping $    \mathcal{T}(\theta^{2*3}_{i}, {G})$}: 
    This warping operation modifies the geometric positions within individual frames based on the transformation parameters \(\theta \in [\theta_{0}, \theta_{1}, \ldots, \theta_{T-1}]\). It recalculates the feature alignment within each frame, adhering to the new spatial coordinates as defined by Eqn.~\ref{eqn:2D}.

    \item \textbf{Temporal Warping $    \mathcal{T}(\theta^{3*4}_{cls}, {G})$}: To capture the video's dynamic nature, Temporal Warping applies $\theta_{cls}$, extending deformations beyond the spatial confines of a single frame into the temporal dimension. This warping models the cross-frame geometric changes, accounting for the motion and temporal evolution of the scene as dictated by Eqn.~\ref{eqn:3d}. 

\end{enumerate}



Given a video feature map \( \tilde{X}_s \in \mathbb{R}^{T \times H \times W \times C} \), we establish a spatial-temporal grid \( G = \{G^o_i\} \) to facilitate the warping process. Each grid element \( G^o_i = (z^o_i, x^o_i, y^o_i) \) corresponds to a feature vector  \( \tilde{X}[G^o_i] \in \mathbb{R}^{C} \).
For each grid element \( G^o_i \) with coordinates \( (z^o_i, x^o_i, y^o_i) \), the warping function \( \mathcal{T} \) computes the new spatial coordinates \( G^s_t \) using \( \theta^{2 \times 3}_t \), and the temporal coordinates \( G^s_{cls} \) using \( \theta^{3 \times 4}_{cls} \). The feature vector at \( G^o_i \) is updated by averaging the contributions from both spatial and temporal mappings. The following equations describe this operation:

\begin{gather}
\label{egn:interpolation}
    G^s_i = \mathcal{T}(\theta^{2 \times 3}_t, G^o_i); \; \;
    G^s_{cls} = \mathcal{T}(\theta^{3 \times 4}_{cls}, G^o_i) \\
    \tilde{X}_o[G^o_i] = \frac{\tilde{X}_s[G^s_i] + \tilde{X}_s[G^s_{cls}]}{2}
\end{gather}

Equipped with the coordinates \( G^s_i \) and \( G^s_{cls} \), we apply differentiable trilinear interpolation to resample input features. This interpolation method selects the nearest neighbors to the computed coordinates \( (z^s_i, x^s_i, y^s_i) \), utilizing four points for spatial warping within the same frame and eight points for temporal warping across frames.

\subsubsection{Comparing  with Vanilla ViT Attention} 

As shown in Figure~\ref{fig:overview}, given the \( i \)-th visual token, such as a pixel or patch, self-attention mechanisms update the token feature \( x_i \) by calculating its similarity to all other tokens. This process, known as Global Attention Aggregation, ensures that each token is updated with a comprehensive understanding of the entire context.

In contrast, Spatial-Temporal Warping adopts a targeted strategy for updating \( x_i \), emphasizing the spatial-temporal geometry inherent to the video sequence. As outlined in Eqn.~\ref{egn:interpolation}, it takes into account specific coordinates \( G^s_i \) and \( G^s_{cls} \) mapped from potential deformations and uses differential interpolation to resample \( x_i \) in conjunction with its nearest coordinates. This strategy adeptly weaves local geometric changes into the feature update mechanism,  endowing visual transformers with the ability to perform geometric deformations on feature maps.

\subsection{Model Architectures} 
\label{sec:arch}

Our model architecture adopts a hierarchical stacking of blocks, a method prevalent in previous action recognition research~\cite{li2022mvitv2}. 
To enhance comparison and demonstrate the efficacy of our approach, we replace the vanilla self-attention blocks with our innovative {\fancyname} blocks within the well-known MViTv2~\cite{li2022mvitv2} action recognition framework.  
As illustrated in Figure~\ref{fig:overview}, the {\fancyname} shares a similar pyramid structure with MViTv2-small, with the number of blocks per stage arranged in a {1, 2, 11, 2} sequence. This structure corresponds to a channel number progression of {96, 192, 384, 768}. Notably, at the commencement of each stage, the video feature map is resized to accommodate the differing demands of the successive stages. Finally, the average pooling and fully connected layer are utilized to output the final predictions.

\section{Experiments}
\label{sec:exp}

\subsection{Dataset and Experimental Setup}
\label{sec:exp_rec}

\noindent\textbf{Datasets.}
We evaluate our method on three action recognition benchmarks, including UCF101~\cite{soomro2012ucf101},
HMDB51~\cite{kuehne2011hmdb},
and Mini-Kinetics-200~\cite{xie2018rethinking}.
The UCF101 and HMDB51 are widely used comprehensive datasets for action recognition. , reflecting a diverse array of human activities.
Mini-Kinetics-200, a condensed version of the extensive Kinetics-400~\cite{kay2017kinetics}, focuses on the 200 most populated categories, providing a dense and varied subset for training and validation.
These video datasets, sourced from an array of internet platforms and digitized movies, encapsulate various real-world challenges, such as camera motion, diverse lighting conditions, and occlusions.

\noindent\textbf{Experimental Setup.} 
We adhere to the training recipe and inference strategies from MViTv2~\cite{li2022mvitv2} for the Kinetics dataset by default.
The strategy for handling the input clip and the spatial domain are identical to those in MViTv2.
Specifically, we use MViTv2-S as the primary baseline for direct comparison, wherein we substitute its transformer blocks with our {\fancyname} blocks to construct the action recognition architecture.
For a comparative analysis, we adopt multi-clip and multi-crop testing for UCF101 and HMDB51. All scores are averaged for the final prediction.
More implementation specifics are shown in the Appendix~\ref{apd:hype_para}.


\noindent\textbf{Evaluation.} 
Our evaluation aims to assess the effectiveness and efficiency of the  {\fancyname} module in improving action recognition performance. 
We report Top-1 and Top-5 accuracy (\%) for Mini-Kinetics-200 and report top-1 accuracy for UCF101 and HMDB51, following the convention of previous works~\cite{wu2021dsanet, carreira2017quo}. 
We run each experiment three times and show the averaged results. Additionally, we measure the computational cost by reporting 
the number of floating-point operations (FLOPs), and the number of parameters (Params). FLOPs are based on processing uniformly sampled $K=16$ clips from a video, each scaled to 224x224 from the original resolution.

\subsection{Main Results}
\label{sec:mainResults}

\begin{table*}[]
\centering
\begin{tabular}{l c c|c c}
    \toprule
     \multirow{2}{*}{\textbf{Models}} & \multicolumn{2}{c|}{\textbf{ Top-1 Accuracy }} & \multicolumn{2}{c}{\textbf{Computation Cost}} \\
    \cline{2-5}
    & UCF101  & HMDB51   &  Params (M) &  FLOPs (G)\\ \midrule
    3D-ConvNet~\cite{tran2015learning} & 51.6  & 24.3  & 79.0 & 108.0  \\ 
    3D-Fused$^*$~\cite{tran2015learning} & 69.5  & 37.7  & 39.0 & - \\
    Two-Stream$^*$~\cite{simonyan2014two} & 83.6 & 47.1  & 12.0 & -\\
    Two-Stream I3D$^*$~\cite{tran2015learning} & 88.8  & 62.2  & 25.0 & 216.0 \\
        \hline 
    MViTv2~\cite{li2022mvitv2} & 85.0  & 69.0 & 34.3 & 64.4   \\

    \textbf{GeoDeformer} &\textbf{89.1} (+4.1)  & \textbf{73.8} (+4.8) &  34.4  & 67.1 \\ 
    \bottomrule
\end{tabular}
\caption{
Experiment results on UCF101 and HMDB51 datasets. We compare recent models with the MViTv2 backbone as well as classic CNN-based methods. The metrics used for comparison include accuracy, the number of parameters (Params, in millions), and computational cost (FLOPs, in billions).  Models marked with $^*$ use optical flow information for action recognition. 
Note that our methods only use these datasets' RGB frames and do not use extra modalities (e.g., optical flow, audio).
The Params and FLOPs values of the classic methods are approximations based on ~\cite{tran2015learning} and ~\cite{feichtenhofer2019slowfast}. FLOPs and Params are reported referencing UCF101, and the values for the HMDB51 dataset are closely similar.}
\label{tab:main_results}
\end{table*}

\begin{table}[]
\begin{center}
\resizebox{1.0\columnwidth}{!}
{
\begin{tabular}{l|c|c|c|c}
\specialrule{0.9pt}{0pt}{0pt}
\textbf{Models} & \textbf{Params}  & \textbf{FLOPs}  &  \textbf{Top-1 }&  \textbf{Top-5}\\
\specialrule{0.9pt}{0pt}{0pt}
\addlinespace[0.5ex]

\specialrule{0.5pt}{0pt}{0pt}
\multicolumn{5}{c}{a) K400-trained} \\
\hline
    I3D+NLN~\cite{wang2018non} & 35.40 & 49.10 & 78.28 & 93.40\\
    SlowFast~\cite{feichtenhofer2019slowfast} & 32.45 & 27.38 & 80.15 & 93.86\\
    CSN-L~\cite{tran2019video} &22.21 & 74.50& 82.60 & 95.56 \\
    R2PLUS1D~\cite{tran2018closer} & 28.11 & 75.67 & 83.25 & 94.91 \\
\specialrule{0.5pt}{0pt}{0pt}
\addlinespace[0.5ex]

\specialrule{0.5pt}{0pt}{0pt}
\multicolumn{5}{c}{b) K200-trained} \\
\hline
    S3D~\cite{xie2018rethinking} & 8.77 & 66.38 & 78.90 & -\\
    I3D~\cite{yue2018compact} & 12.06 & 107.89 & 77.40 & 93.20\\
    CGNL~\cite{yue2018compact} & - & - & 79.90 & 93.40\\
    V4D~\cite{zhang2020v4d} & - & 143.00 & 80.70  & 95.30 \\
    DSA~\cite{wu2021dsanet} & - & 83.80 & 81.80 & 95.40\\
\hline
  MViTv2~\cite{li2022mvitv2} & 34.31 & 64.39  & 82.23  & 95.71  \\
  \textbf{GeoDeformer}  &  34.56 & 67.12 & \textbf{83.38}  & \textbf{96.57} \\ 
\specialrule{0.5pt}{0pt}{0pt}
\end{tabular}
}
\end{center}
\caption{
Comparative results on the test set of Mini-Kinetics-200~\cite{xie2018rethinking}. The table presents the performance of various models, with a focus on the top-1 and top-5 accuracy metrics. Models listed in b) are trained from scratch on K200, while those in a) are trained on the Kinetics-400 but tested on the K200 subset. } 
\label{tab:main_results_k200_v2}
\end{table}



We evaluate the effectiveness of our proposed {\fancyname} module on the UCF101~\cite{soomro2012ucf101},
HMDB51~\cite{kuehne2011hmdb},
and Mini-Kinetics-200~\cite{xie2018rethinking}. The primary results are presented in Table~\ref{tab:main_results} and Table~\ref{tab:main_results_k200_v2}.

\noindent\textbf{UCF101 and HMDB51.} Table~\ref{tab:main_results} highlights the {\fancyname}'s capacity to significantly boost action recognition accuracy in both UCF101 and HMDB51 datasets. 
On UCF101, only substituting one of the MViTv2 blocks with our {\fancyname} improves the accuracy by an absolute 4.1\% in top-1 accuracy, with a minimal additional computational cost. For HMDB51, the enhancement is even more pronounced. The {\fancyname} block leads to a 4.8\% improvement in top-1 accuracy compared to the MViTv2 model. This highlights the critical role of spatial-temporal geometric transformations in action recognition, especially in relatively smaller datasets.


\noindent\textbf{Mini-Kinetics-200.} Following previous works~\cite{xie2018rethinking, zhang2020v4d, wu2021dsanet}, Mini-Kinetics-200 (K200), a subset of the Kinetics-400 dataset, includes 200 categories with the most training instances. 
It serves as a testbed to evaluate the efficacy of our {\fancyname} on larger datasets.
We are not able to use the full Kinetics-400 dataset due to limited computational resources.
As the results listed in Table~\ref{tab:main_results_k200_v2} (b), 
our approach has demonstrated significant enhancements in the action recognition task with a 1.15\% increase in Top-1 accuracy, establishing a new state-of-the-art on this dataset.



\subsection{Ablation Experiments}
\label{sec:ablation}

To verify the efficacy of the {\fancyname} and its components, we undertake a series of ablation studies on the UCF101 dataset. These studies aim to refine and substantiate the foundational insights that inform {\fancyname}'s design. 
We investigate the following questions:

\begin{table}[t!]
\centering
\begin{tabular}{l|c|c}
\toprule
Method
& {\em }Top-1{\em } & {\em }FLOPs{\em }  
               \\

\specialrule{0.9pt}{0pt}{0pt}
\addlinespace[0.4ex]

\specialrule{0.4pt}{0pt}{0pt}
MViTv2-Baseline              & 84.97 &   64.39     \\ 
\hline

\multicolumn{3}{c}{ \em (a) Spatial-Temporal Deformations } \\
Spatial Only  & 88.42 & 67.11 \\
Temporal  Only  & 87.34 & 67.11 \\
Spatial $\&$ Temporal  & 89.08 & 67.12 \\
\midrule

\multicolumn{3}{c}{ \em (b) Transformation Type Constraints } \\

Focusing Transformation     & 87.86  & 67.12  \\
Affine Transformation & 89.08 & 67.12 \\
Homography Transformation  & 89.16 & 67.12 \\
\midrule

\multicolumn{3}{c}{ \em (c) Deformation Predictor Designs } \\

Without Attention Matrix & 87.95 & 66.70  \\
Without Temporal Attention  & 86.95 & 65.00  \\
{GeoDeformer} & 89.08 & 67.12 \\
\midrule

\multicolumn{3}{c}{ \em  (d)  Model Architecture Optimazations} \\

{GeoDeformer}$[0]$ & 89.08 & 67.12 \\
{GeoDeformer}$[1]$ & 88.83 & 67.11 \\
{GeoDeformer}$[3]$ & 88.23 & 67.10 \\
{GeoDeformer}$[14]$ & 87.89 & 65.51 \\
{{GeoDeformer}$[0,1,2]$ }& 88.98 & 73.60 \\ 
{GeoDeformer}$[0,1,3,14] $ & 88.81 & 76.33 \\
\midrule

\multicolumn{3}{c}{ \em (e) Cross Dataset Transferability } \\


HMDB51 \text{--$>$} UCF101  & 87.89 & 67.12  \\
K200 \text{--$>$} UCF101  & 88.24 & 67.12  \\
UCF101 \text{--$>$} K200  & 82.84 & 67.12  \\


\bottomrule
\end{tabular}

\caption{Ablation experiment results on UCF101, consistent with methodologies detailed in Section \ref{sec:exp_rec}. (a) Examining the impact of in-frame spatial and cross-frame temporal deformations. (b) Assessing constraints in geometric deformation parameterization. (c) Evaluating the efficacy of the geometric deformation predictor module $\mathcal{D}$. (d) Identifying optimal strategies for constructing action recognition frameworks with {\fancyname}. (e) Investigating the transferability of the {\fancyname} block across different datasets.}
\vspace{-4mm}
\label{tab:ablation}
\end{table}

\vspace{-2mm}
\subsubsection{Do spatial and temporal deformation help?}

Building upon the {\fancyname}'s dual capability of spatial and temporal geometric transformations (\cref{sec:overview}), this investigation examines their individual and combined effects on action recognition performance. The empirical analysis, as presented in Table~\ref{tab:ablation} (a), reveals that both spatial and temporal deformations contribute positively to recognition accuracy. More importantly, when both deformations are integrated, they amplify the model's performance, underscoring the value of their confluence in enhancing action recognition.

\vspace{-2mm}
\subsubsection{What is the best transformation constraint?}

In Section~\ref{sec:stdlm}, we introduced geometric transformations that employ parameterization matrices, 
These transformations, i.e., $\Theta$, are referred to as affine transformations (see Eqn.\ref{eqn:2D} and Eqn.~\ref{eqn:3d}), denoted as \(\mathcal{T}\). 
In this section, we expand our discussion to include additional parameterization strategies: the Focusing Transformation
($\mathcal{T}_F$) and the Homography Transformation ($\mathcal{T}_H$).

The Focusing Transformation
($\mathcal{T}_F$) is more restrictive, where operations such as cropping, translation, and scaling are permitted. For instance, for $\theta_{cls}$, the transformation includes six learnable parameters and is defined as:

\vspace{-2mm}
\begin{align}
\label{eqn:focusing}
\mathcal{T}_F(\theta, G):
\begin{bmatrix}
Z^s \\
X^s \\
Y^s \\
\end{bmatrix}
=
\begin{bmatrix}
p_0 & 0 & 0 & p_3\\
0 & p_1 & 0 & p_4\\
0 & 0 & p_2 & p_5\\
\end{bmatrix}
\begin{bmatrix}
Z^o \\
X^o \\
Y^o \\
1 \\
\end{bmatrix}
\end{align}

In contrast, the Homography Transformation allows for non-homogeneous transformations, which have a higher degree of freedom (DoF), as demonstrated in the following equation:

\begin{align}
\vspace{-4mm}
\label{eqn:3d_homography}
\mathcal{T}_H(\theta, G):
\begin{bmatrix}
Z^s \\
X^s \\
Y^s \\
M \\
\end{bmatrix}
=
\begin{bmatrix}
p_1 & p_2 & p_3 & p_4\\
p_5 & p_6 & p_7 & p_8\\
p_9 & p_{10} & p_{11} & p_{12}\\
p_{13} & p_{14} & p_{15} & 1 \\
\end{bmatrix}
\begin{bmatrix}
Z^o \\
X^o \\
Y^o \\
1 \\
\end{bmatrix}
\end{align}

\noindent where the feature coordinates $(z^o_i,x^o_i,y^o_i)$ are mapped to the sample feature coordinates $(z^s_i/m_i,x^s_i/m_i,y^s_i/m_i)$, incorporating a projective component through $M$, which allows for perspective transformations.

As indicated in the results listed in Table~\ref{tab:ablation} (b), compared to affine transformations, the focusing transformations are less effective, while the performance with homogenous transformations shows a slight improvement. This observation suggests that a higher degree of freedom in the transformation parameterization leads to higher performance. However, considering that the homogenous transformation only results in a marginal improvement of 0.08 and potentially reduces the interpretability of the model, we opt for affine transformations as our default setting.

\vspace{-2mm}
\subsubsection{What is a good deformation predictor design?}

As detailed in Section~\ref{sec:stdlm}, our methodology for geometric deformation detection incorporates two primary design strategies to enhance the learning of transformation parameters: Global Attention Matric Enhance and Temporal Attention Integration. 
We examine the impact of these two designs.

As presented in Table~\ref{tab:ablation} (c), 
When only the raw frame-wise feature map is used for geometric deformation detection, the top-1 accuracy drops to 87.95. 
Meanwhile, without the temporal self-attention mechanism, the performance decreases to 86.95. 

\vspace{-2mm}
\subsubsection{What is the optimal model architecture?}
The standard ViT action recognition frameworks, exemplified by MViTv2~\cite{li2022mvitv2}, are characterized by a multi-stage structure where each stage comprises several attention blocks. 
In our study, we integrate our {\fancyname} block into MViTv2 by replacing the standard MViT blocks. 
Considering the MViTv2-small architecture, the distribution of blocks across the stages follows a {1, 2, 11, 2} pattern, totaling 16 blocks. We denote the replacement of the first, second, fourth, and fifteen blocks in MViTv2 with our {\fancyname} block as {\fancyname}$[0,1,3,14]$, which addresses replacing one of the blocks at the onset of each stage. {\fancyname}$[0, 1, 2]$ indicates that all the blocks of the first and second stages are replaced as {\fancyname} Block.

Our findings, as depicted in Table~\ref{tab:ablation} (d), reveal that replacing shallower blocks tends to yield better performance, albeit with a slight increase in computational cost. 
Specifically, {\fancyname}$[0]$ achieved a Top-1 accuracy of 89.08, while {\fancyname}$[14]$ registered an accuracy of 87.89.
Surprisingly, we observe that adding multiple {\fancyname} blocks does not lead to performance enhancement. The most effective result is obtained by implementing a single geometric deformation at the very beginning of the network.


\subsubsection{Can {\fancyname} transfer to a new dataset?}

To evaluate the {\fancyname} model's transfer learning capability, we examine its performance on different datasets with fixed initialization parameters of {\fancyname}[0].
Table~\ref{tab:ablation} (e) presents our findings, which include transferring parameters from HMDB51 and Mini-Kinetics-200 to UCF101. Results indicate that the {\fancyname} model, when initialized with parameters from both smaller (HMDB51) and larger (K200) datasets, surpasses the baseline performance on UCF101 (87.89 vs. 84.97 and 88.24 vs. 84.97 at top-1). Additionally, transferring parameters from UCF101 to K200 also yields better results than training from scratch on K200 (82.84 vs. 82.23 at top-1).
This suggests that the deformation learned from one dataset can indeed be effectively transferred to another, highlighting the potential of {\fancyname} for cross-dataset action recognition tasks.



\subsection{Qualitative Analysis}
\label{sec:visualize}

In this section, we visualize example transformations of our model to illustrate the deformation effects.

\begin{figure}[t!]
    \vspace{-6mm}
	\centering
		\includegraphics[width=0.47\textwidth]{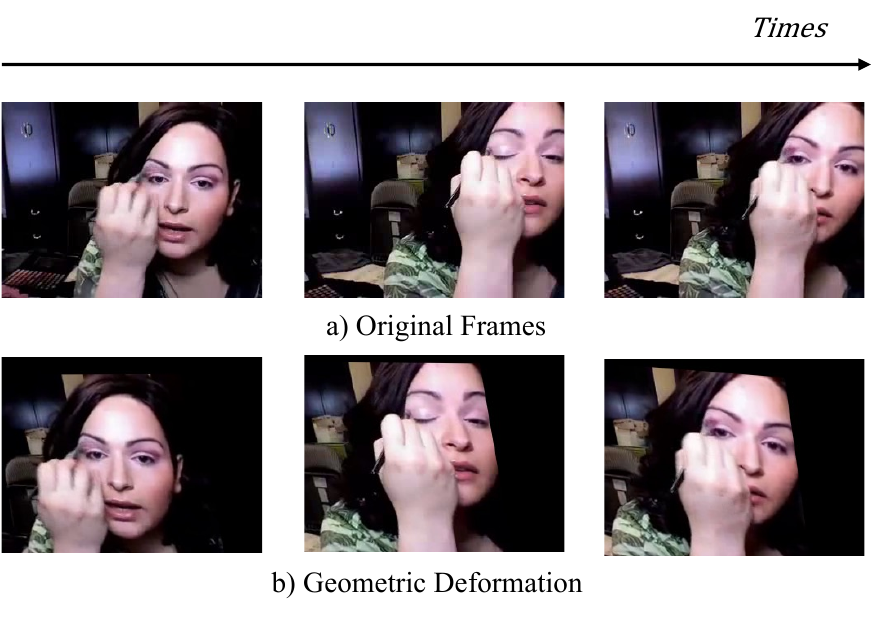} 
	\caption{Qualitative analysis. This figure displays a sequence of video frames before and after applying the {\fancyname} model's transformation. The top row depicts original frames from a video where a woman is applying makeup, initially centered but then moving to the right side of the frame. The bottom row shows the transformed frames, where the model has adjusted the positioning and scale to re-center the subject. 
	}
    \vspace{-4mm}
    \label{fig:qual}
\end{figure}

We apply the per-frame transformation (\(\theta\), as per Eqn. \ref{eqn:3d}) from the {\fancyname} block, to the original video sequences and focus on illustrating the spatial transformation of the corresponding middle three frames. 
As depicted in Fig. \ref{fig:qual}, the first row presents the original video frames, capturing a woman engaging in the act of applying makeup. 
Initially centered, she subsequently moves to the right, drifting away from the center of the frame. The spatial deformation learned by our model effectively centralizes the actor. More visualization results can be found in the Appendix~\ref{apd:hype_case}.


\section{Conclusion}

This paper introduces the Geometric Deformable Transformer (\textit{\fancyname}), a novel approach that integrates explicit geometric understanding into the transformer architecture for action recognition. 
The \textit{\fancyname} model excels at identifying spatial and temporal geometric variations in visual data, thus adaptively adjusting both within-frame geometries and cross-frame dynamics for better performance. 
Our comprehensive experiments confirm \textit{\fancyname}'s superiority over established baselines, showcasing its potential to enhance visual transformer capabilities. 


{
    \small
    \bibliographystyle{ieeenat_fullname}
    \bibliography{main,customer}
}


\newpage

\appendix

\newpage
\clearpage


\section{Hyper-parameters of Geometric Deformation Predictor}
\label{apd:hype_para}

As discussed in \cref{sec:stdlm}, the Geometric Deformation Predictor \(\mathcal{D}\) is responsible for learning deformation parameters \(\Theta\) to effectively model both spatial and temporal geometric deformations within each frame and across the entire video. By integrating self-attention mechanisms with CNN components, \(\mathcal{D}\) adeptly manages spatial-temporal deformations in videos. This results in a unified representation that encapsulates the intricacies of spatial and temporal changes.

The hyper-parameters of the Geometric Deformation Predictor, as used in this work, are detailed in Table~\ref{tab:hype_para}.

\begin{table}[h!] 
    \centering
     \setlength{\tabcolsep}{1pt} 
    \begin{tabular}{l c c c}
        \toprule
        \multirow{2}{*}{\bf Layers} & \multirow{2}{*}{\bf Parameter} &  \multirow{2}{*}{\bf  Channel} \\
        \\
        \midrule
         2D-CNN-layer1  & k=3, s=2  &768 \\
         2D-CNN-layer2  & k=3, s=2  & 192 \\
         AdaptiveAvgPool2d & - & 192 \\
         attention-layer1 & heads=4 & 192 \\
         activation function & gelu & - \\
         attention-layer2 & heads=4 & 192 \\
         activation function & gelu & - \\
         FC-1 & - & 12  \\
         FC-2 & - & 6  \\
        \bottomrule
    \end{tabular}
    \caption{Hyperparameters of Geometric Deformation Predictor.}
    \label{tab:hype_para}
\end{table}

In our proposed method, the initial stage involves employing a 2D Convolutional Neural Network (CNN), followed by an Adaptive Average Pooling layer (AdaptiveAvgPool2d). This combination is utilized to extract frame embeddings, which are denoted as \( H = \{h_{\text{cls}}, h_0, h_1, \ldots, h_{K-1}\} \). These embeddings are then fed through two attention layers to enhance the feature representation. The final stage of the process involves fully connected layers, which transform the embeddings \( H \) into geometric transformer parameters, specifically \(\theta_t \in \mathbf{R}^{2 \times 3}\) and \(\theta_{cls} \in \mathbf{R}^{3 \times 4}\).

\section{Extended Case Studies}
\label{apd:hype_case}

In this Section, we extend our qualitative analysis by presenting additional cases that further demonstrate the efficacy and versatility of the {\fancyname} model in handling diverse scenarios. These cases showcase the model's ability to adapt to various video contexts and emphasize the practical applicability of the spatial and temporal deformations learned by the model. Through these examples, we aim to provide a broader understanding of the model's performance and its impact on video frame transformation.


\begin{figure}[ht]
    \centering        
    \includegraphics[width=0.48\textwidth]{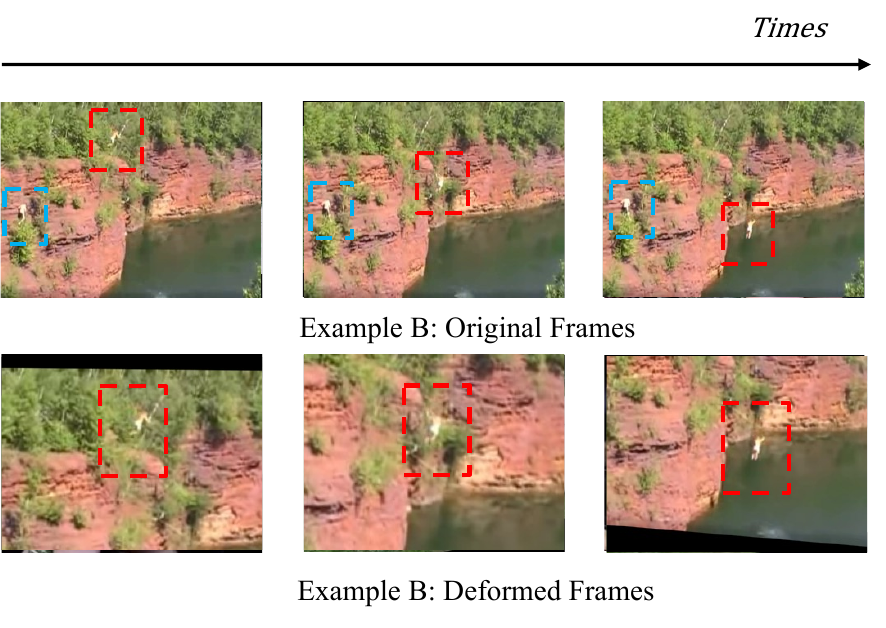} 
    \caption{Example B shows a sequence of video frames before and after the application of the geometric deformation by the {\fancyname} model. In the top row, the original frames depict an actor cliff diving, shot with a narrow field of view and the actor initially positioned at the edge of the frame. The transformation applied by {\fancyname} not only centers the main actor but also brings the field of view closer to enhance focus on the central action. Importantly, it also effectively crops out irrelevant characters on the left, streamlining the frame to emphasize the primary subject of the scene. The bounding boxes are added manually for illustration purposes.}
\end{figure}

\begin{figure}[ht]
    \centering		
    \includegraphics[width=0.48\textwidth]{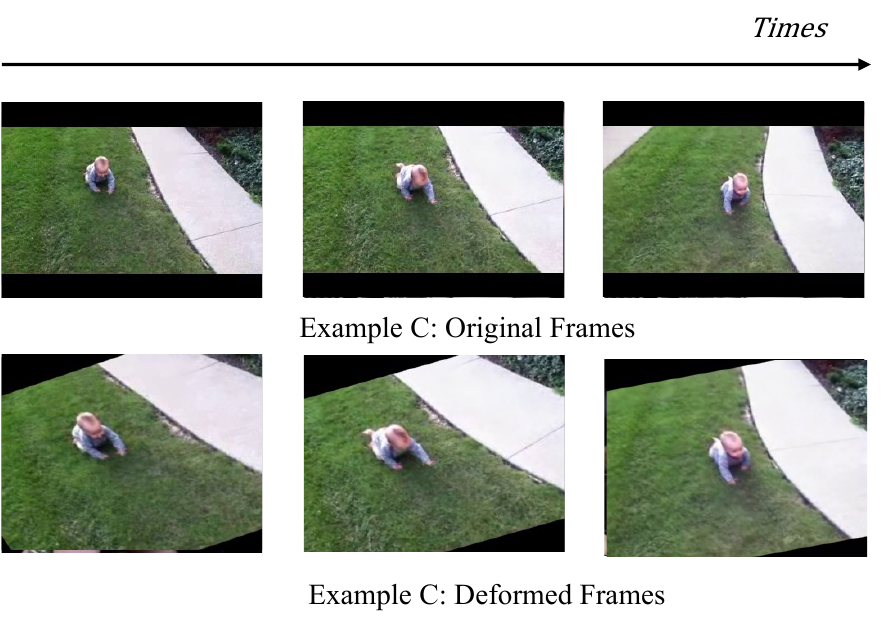} 
    \caption{Example C displays a sequence of video frames before and after applying the {\fancyname} deformation. The top row depicts original frames from a video where a baby is crawling, captured from a unique camera angle with a top-down perspective. The {\fancyname} model effectively learns a transformation that not only magnifies the main actor, the baby, but also alters the video's perspective, leading to a view that emphasizes the baby’s rightward movement.}
\end{figure}

\label{sec:appendix}

\end{document}